# AGENTIC AI FOR AUTONOMOUS ANOMALY MANAGEMENT IN COMPLEX SYSTEMS

Reza Vatankhah Barenji [1*], Sina Khoshgoftar[1]

Department of Engineering, School of Science and Technology, Nottingham Trent University, Nottingham, NG118NS, UK

Reza.vatankhahbarenji@ntu.ac.uk

**Abstract**

This paper explores the potential of Agentic AI in autonomously detecting and responding to anomalies within complex systems, emphasizing its ability to transform traditional, human-dependent anomaly management methods. Building on recent advancements, the study illustrates how Agentic AI—AI agent augmented with large language models, diverse tools, and knowledge-based systems—continuously analyses and learns from vast, multi-source datasets to autonomously identify, interpret, and respond to abnormal behaviours in complex, adaptive systems. Unlike conventional AI agents constrained by predefined roles, Agentic AI synthesizes insights across disciplines, detects subtle patterns, and adapts its strategies using both implicit and explicit knowledge. This paper underscores the need to evolve current human-based anomaly management approaches toward fully autonomous systems, highlighting Agentic AI's adaptive, goal-driven nature—particularly in complex contexts where traditional methods often fall short.

## 1. Introduction

A complex system operates through the continuous interactions among digital infrastructure, organizational stakeholders, processes, and policy or rule-based governance. These systems are inherently complex due to environmental uncertainties and the intricate interdependencies among their components and actors, often functioning across multiple operational modes many of which are rare or unpredictable (Y. Lin et al., 2025). Anomalies in such systems may arise from both internal and external disturbances, potentially disrupting functionality, compromising reliability, or reducing overall efficiency(Costantino et al., 2022; Y. Fang et al., 2020). Internally, disruptions may result from sensor malfunctions, human error, workflow inefficiencies, or component failures, leading to delays, performance degradation, and increased operational costs. Externally, factors such as regulatory changes, resource shortages, and environmental events can also negatively impact performance, causing downtime, dissatisfaction among users, and financial losses.

Since anomalies often indicate potential risks, anomaly management is essential for understanding the nature and causes of underlying issues and effectively addressing them (Y. Fang et al., 2020; Z. Li et al., 2022). Anomaly management involves both the diagnosis and resolution of unexpected system behaviours through appropriate interventions aimed at restoring or maintaining system functionality—tasks that are traditionally performed by human experts (Costantino et al., 2022). Diagnosis refers to the analysis of the nature or cause of a condition, situation, or problem based on specific signs, patterns, or symptoms (Z. Li et al., 2022).

Artificial Intelligence (AI) plays a significant role in anomaly diagnosis by offering insights, simulating scenarios, and providing predictive recommendations. By rapidly processing large datasets, AI reduces cognitive load and enables faster, more adaptive responses in anomaly management. Unsupervised learning and deep learning methods, in particular, have proven effective for anomaly detection and interpretation (Mattera et al., 2025). Despite these advancements, anomaly management in complex systems still heavily relies on human decision-making, as many operations require human oversight, judgment, and accountability (Ding et al., 2023). Once an anomaly is diagnosed, human experts are typically responsible for making decisions and applying interventions to mitigate or resolve the abnormal condition.

Agentic AI, representing the fourth wave of artificial intelligence, builds upon traditional AI agents that augmented by deep learning (DL) methods such as large language models (LLMs). This enables autonomous decision-making, contextual understanding, and the capacity to perform complex, multi-step tasks with minimal human intervention (Acharya et al., 2024; Chen & Chan, 2024; Estrada-Torres et al., 2024; C. Fang et al., 2024; Yuksel & Sawaf, 2024). Agentic AI holds the potential not only to replicate but also to surpass human capabilities in anomaly management within complex systems. These systems can autonomously carry out reasoning, planning, and action in real time (Zhuge et al., 2024).

Agentic AI may significantly improve anomaly management in complex systems by autonomously analyzing and learning from vast, multi-source datasets, while leveraging a diverse set of tools to adapt to evolving conditions and enhance decision-making over time using both implicit and explicit knowledge of the system (Kapoor et al., 2024; Zhuge et al., 2024). Unlike human operators in conventional anomaly management systems—who are often constrained by domain specialization and cognitive biases—Agentic AI can synthesize insights across multiple disciplines, detect subtle or hidden patterns, anticipate anomalies, and support high-precision decision-making. Its ability to make objective and consistent decisions at scale allows Agentic AI to potentially outperform human counterparts in terms of speed, scalability, and accuracy.

This study is motivated by the limitations of current human-based anomaly management approaches and explores how Agentic AI can autonomously manage anomalies in complex systems. The research questions guiding this study are:

RQ1: How do traditional AI methods perform in anomaly management?

RQ2: What limitations exist in current anomaly management within complex systems?

RQ3: What capabilities distinguish Agentic AI from conventional AI agent in terms of autonomy, adaptability, and long-term goal management?

RQ4: How can Agentic AI support real-time anomaly detection, interpretation and intervention in complex system?

This study adopts a narrative literature review methodology to examine the role of Agentic AI in autonomous anomaly management within complex systems (Rother, 2007). The review analysed 89 articles directly addressing Agentic AI, primarily within the fields of artificial intelligence, computer science, and machine learning. Additionally, over 52 papers were reviewed on anomaly management, novelty detection, and novelty interpretation. This interdisciplinary synthesis forms the conceptual foundation for applying Agentic AI to complex systems.

The paper is structured as follows. First, we discuss anomalies and their management in complex systems, along with the role of AI in this context. This section concludes by identifying the shortcomings of current approaches to anomaly management in complex systems. We then examine the evolution of AI, with a particular focus on AI agents and the emergence of Agentic AI—exploring their functionality, development, and the key differences. Next, we investigate how Agentic AI enhances anomaly management in complex systems, emphasizing its applications and benefits. Following this, we discussed the broader implications of Agentic AI for anomaly management and its impact on the resilience and adaptability of complex systems. Finally, we provide concluding remarks.

## 2. Exploring Anomalies and AI's Contribution to Their Management

### 2.1 Anomaly definition

Anomalies also known as outliers, abnormalities, or deviants are data patterns that deviate significantly from expected behaviour. They are generally defined as data observations that substantially diverge from a defined concept of normality. (Prasad et al., 2009) characterized anomalies as patterns in data that do not conform to a well-defined notion of normal behaviour. Similarly, (Pang et al., 2021) described anomalies as data instances that significantly deviate from the majority of data instances. Given that data in complex systems often take the form of time series or spatiotemporal modalities (e.g., images or videos), (Erhan et al., 2021) identified anomalies as unusual local changes in spatial or temporal values. Two foundational principles help define anomalies: (a) an anomaly is a relative concept, defined in contrast to a given notion of normality, and (b) an anomaly involves a considerable deviation from this notion of normality.

(Ruff et al., 2021) defined an anomaly as an observation that deviates considerably from a concept of normality, formalized using probability theory. They emphasized that a core challenge in anomaly detection is the unsupervised nature of the problem there is often no explicit definition of normal or abnormal behaviour in most applications. As a result, assumptions must be made regarding the nature of normality or abnormality based on the specific application, domain, and available data. These assumptions, typically developed empirically, are critical for handling the unsupervised nature of anomaly detection and offer valuable insights into designing new detection approaches.

Anomalies can be classified across multiple dimensions based on their characteristics and origins (Samariya & Thakkar, 2023). From a pattern recognition perspective, anomalies are generally categorized into three main types: (a) point anomalies: individual data points that deviate significantly from the norm, (b) contextual anomalies: data points that are considered abnormal only under specific contextual conditions such as time, location, or system state, and (c) collective anomalies: groups of data points that, while individually appearing normal, collectively form an anomalous pattern. Figure 1 presents a schematic of the three main types.

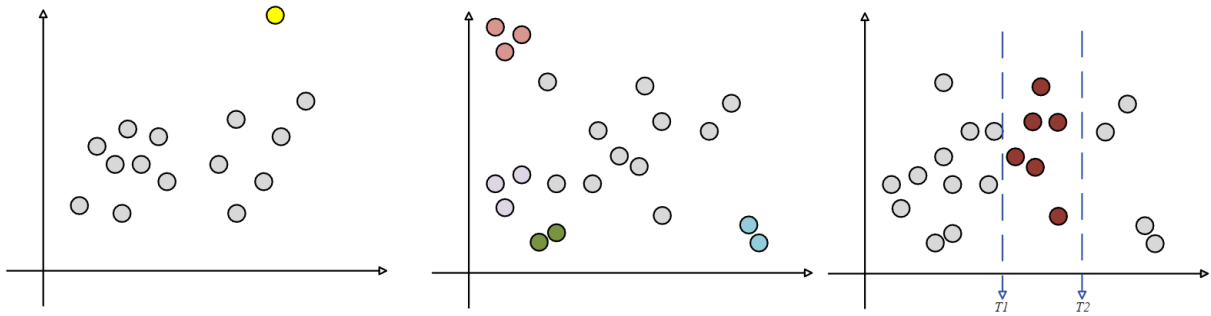

Figure 1 Classification of anomalies based on their characteristics: from left to right point anomaly, group anomaly, context anomaly

In complex systems, anomalies may arise from a variety of sources (shown in Figure 2). These may include system design and modelling errors, human or environmental factors, communication and data issues, or failures within internal or external systems. These anomalies introduce diverse and often subtle patterns into the data, complicating both detection and interpretation. At the data level, anomalies typically manifest as spikes, persistent noise, constant values, or gradual sensor drift. These phenomena can degrade system accuracy and responsiveness at the semantic level, anomalies can be further classified into, (a) low-level anomalies: errors such as sensor faults or image distortions, and (b) High-level anomalies: conceptual misinterpretations resulting from unusual but valid combinations of known inputs (Sejr & Schneider-Kamp, 2021). For example, an autonomous vehicle might misinterpret inactive traffic lights being transported on another vehicle or respond incorrectly to a billboard displaying a stop sign. These classifications underscore the diverse and multifaceted nature of anomalies in complex systems. Addressing such variability requires sophisticated detection strategies capable of identifying both surface-level irregularities and deep contextual or semantic anomalies.

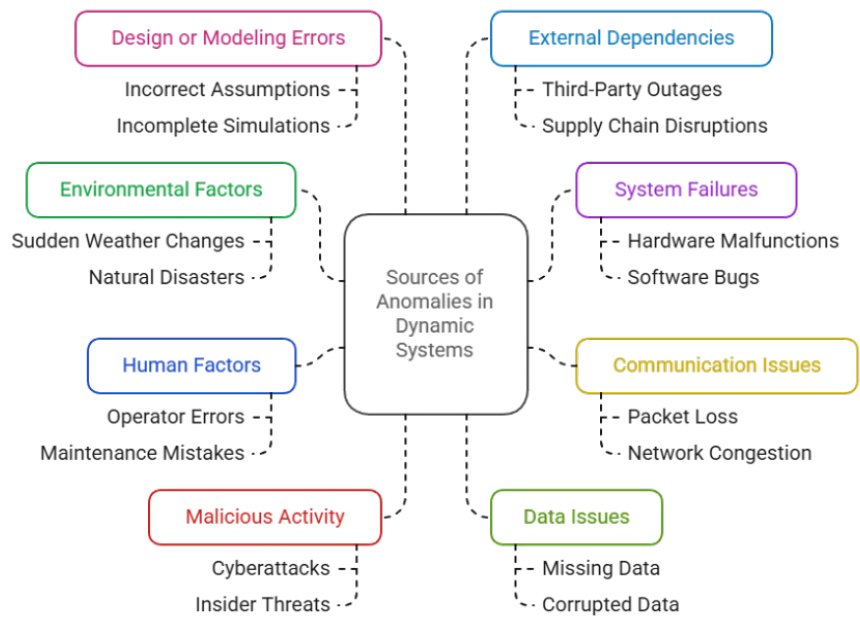

*Figure 2. Classification of anomalies based on their sources*

## 2.2 Anomaly management

Anomaly management is a process that encompasses two main stages: diagnosis and intervention. The diagnosis phase typically involves two key steps—anomaly detection and interpretation. This stage begins with the identification of anomalies, followed by an analysis of the reasons and sources behind the occurrence. Anomaly detection refers to the identification of data points, patterns, or behaviors that significantly deviate from the norm, often signalling potential faults, risks, or unusual events. Once detected, anomaly interpretation aims to understand the underlying causes and contextual relevance of the anomaly. This involves the use of historical data, domain expertise, and pattern recognition to differentiate between harmless outliers and critical issues. This interpretive step is crucial for minimizing false positives and prioritizing appropriate response efforts (Sejr & Schneider-Kamp, 2021).

The second stage, intervention, involves formulating and implementing appropriate actions based on the diagnosis. In complex systems, this stage is typically carried out by human experts who assess the nature of the diagnosed anomaly and determine the most appropriate response. As a result, the human operator remains central to the intervention process. While various tools and technologies may assist with diagnosis and offer decision-support insights, it is ultimately the human expert who makes the critical decisions regarding both the type and timing of the intervention.

Figure 3 summarizes various approaches that can be used for anomaly detection, interpretation, and intervention. Artificial Intelligence (AI) has demonstrated promising capabilities in identifying and understanding anomalies. However, in complex anomaly management systems, the intervention stage remains heavily reliant on human expert decision-making and is neither autonomous nor fully automated. Although some efforts have been made to automate aspects of anomaly management, the majority of current applications are still confined to detection and interpretation, with limited integration into the action or response phase.

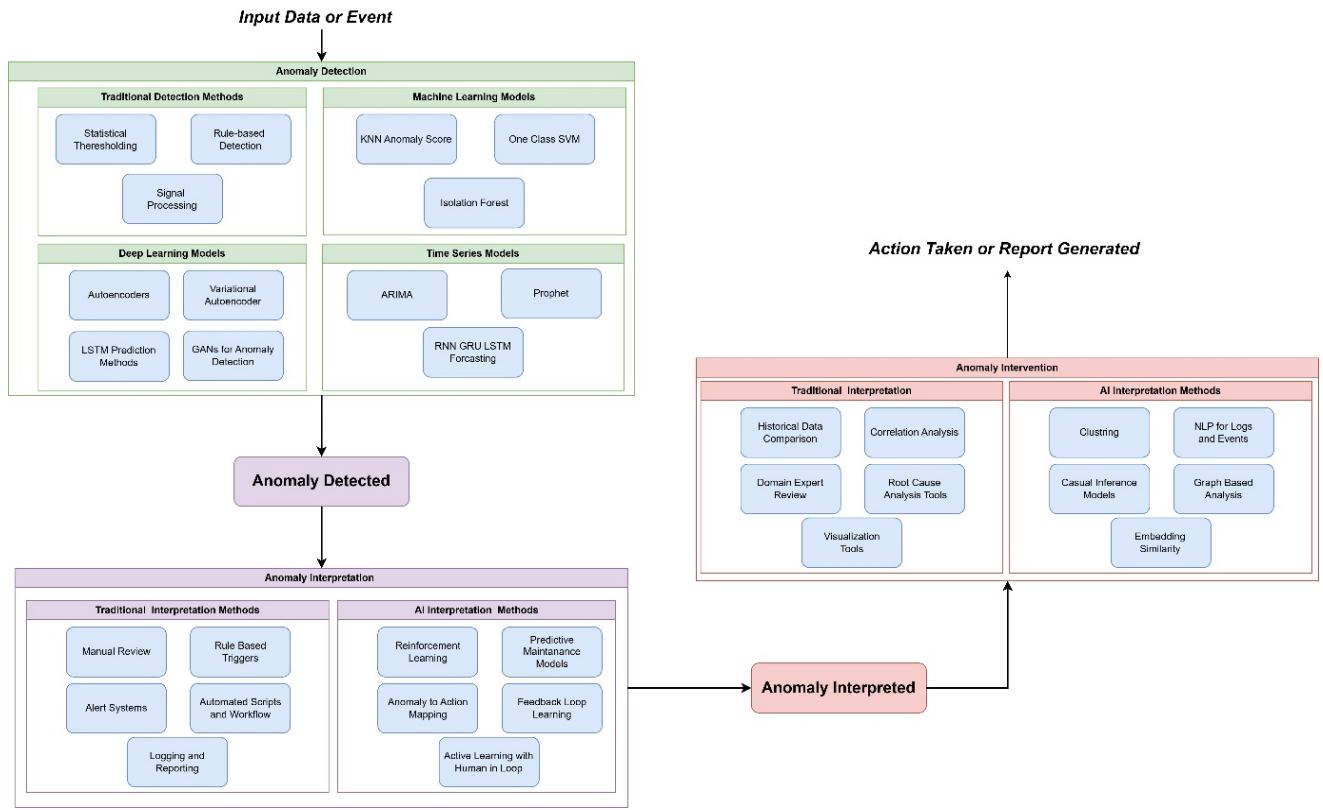

Figure 3. Approached used for anomaly management

## 2.3 AI role in Anomaly detection

The techniques used for anomaly detection can be classified as traditional and AI based methods. Statistical approaches are among the earliest and most widely used methods. These techniques operate on the assumption that normal data follows a specific statistical distribution, with anomalies detected as deviations from this norm. Parametric models rely on estimating parameters of a known distribution, such as Gaussian or regression models (Lee & Chen, 2025). In contrast, non-parametric models make no assumptions about the underlying data distribution and derive structure directly from the data using methods like histograms or kernel density estimation (Moreo et al., 2025). Semi-parametric models incorporate features of both approaches, applying localized modelling strategies to overcome the limitations of global distribution assumptions (Mahmoud, 2021). Although these statistical methods are appreciated for their transparency and interpretability particularly in safety-critical domains they often underperform when handling high-dimensional datasets or when the data distribution diverges from expected patterns.

In environments where data is generated sequentially, such as sensor-rich or real-time systems, time series analysis becomes especially relevant (Xu et al., n.d.). These methods focus on modelling temporal dependencies and forecasting future values based on historical observations (Yao, 2025). For instance, ARMA and ARIMA models are employed to identify trends and patterns in stationary and non-stationary data. Cross-correlation analysis is useful for exploring relationships between multiple time series, and Kalman filtering is widely used for real-time estimation of system states from noisy measurements (Lan et al., 2025). Additional approaches such as autoregressive modelling, symbolic time series analysis, and seasonal decomposition have shown effectiveness in capturing complex temporal patterns and identifying deviations indicative of anomalies (Naveen Kumar et al., 2025).

Signal processing techniques also offer proven solutions for detecting anomalies, particularly when the data is noisy or requires transformation to reveal latent patterns (Alamr & Artoli, 2023). Fourier transforms facilitate the decomposition of time-domain signals into frequency components, making it possible to detect periodic or frequency-based anomalies (Xie et al., 2024).Wavelet transforms build on this by providing a time-frequency representation that is especially effective in detecting both transient and persistent anomalies (Yao et al., 2023). Spectral methods such as Principal Component Analysis (PCA) project high-dimensional data into a subspace that captures the most significant variance, allowing deviations from dominant trends to be more easily identified (Huang et al., 2021). These methods are particularly useful in preprocessing or highlighting anomalies in complex, high-dimensional datasets.

Information-theoretic approaches present another perspective by analysing the informational characteristics of data. These methods use metrics such as entropy, relative entropy, and Kolmogorov complexity to detect disruptions in the expected informational structure of datasets (Q. Zhou et al., 2024). Unlike statistical methods, information-theoretic approaches do not require assumptions about data distribution, making them well-suited to unsupervised learning contexts (Ye et al., 2021). However, their effectiveness can be constrained when anomalous instances are rare or subtle, as the resulting disturbance in information content may not be pronounced enough to trigger detection thresholds.

In AI domain, anomaly detection is commonly framed as a classification problem, where models are trained to distinguish between normal and anomalous patterns. Machine learning-based approaches to anomaly detection can be further categorized based on the learning paradigm used to define normality (Kamalov et al., 2021). Supervised learning can offer high detection accuracy and rapid response times, but its application is limited by the challenge of obtaining comprehensive labels for all types of anomalies, which is often impractical in real-world scenarios (Z. Lin et al., 2022). Consequently, unsupervised learning has gained considerable attention. These methods operate under the assumption that the majority of training data represents normal behaviour, enabling models to flag test instances that differ significantly from the learned representation as anomalies.

Semi-supervised learning, which leverages a small amount of labelled data (often including labelled anomalies), is also used to enhance performance and increase model robustness. Despite their benefits, these approaches are susceptible to overfitting due to the limited availability of labelled anomalies (Memarzadeh et al., 2022). In response to this challenge, self-supervised learning—a subset of unsupervised learning (Gui et al., 2024) has become increasingly popular. Self-supervised methods define auxiliary tasks, known as pretext tasks, that guide the model to learn useful representations from unlabelled data. These learned representations are then applied to downstream tasks such as anomaly detection, often resulting in improved performance (C.-L. Li et al., 2021) Reinforcement learning has also been explored in contexts where continuous interaction with the environment allows the model to learn effective strategies for anomaly detection and response (Müller et al., 2022).

Recent advances in deep learning have significantly improved the effectiveness of anomaly detection, particularly in large-scale, high-dimensional, and unstructured data environments. Convolutional Neural Networks (CNNs) excel at capturing hierarchical spatial features, making them especially suitable for tasks involving image data or inputs from multiple sensors (Khan et al., 2022). Autoencoders, which compress input data into a latent space and then attempt to reconstruct it, detect anomalies by measuring reconstruction errors—discrepancies between the input and the reconstructed output signal potential deviations from normal behaviour

Recurrent Neural Networks (RNNs), and more specifically Long Short-Term Memory (LSTM) networks, are well-suited for time-series data, as they can model long-range temporal dependencies and capture sequential patterns (Mienye et al., 2024). Generative Adversarial Networks (GANs), which consist of a generator that creates synthetic data and a discriminator that evaluates its authenticity, detect anomalies based on deviations from the learned data distribution. Probabilistic models such as Restricted Boltzmann Machines (RBMs) and Deep Belief Networks (DBNs) offer additional capacity for modelling complex and uncertain data structures (Luo et al., 2022). Furthermore, hybrid models that combine deep learning architectures—such as autoencoder-CNN or LSTM-GAN frameworks—often outperform single-model approaches by leveraging the complementary strengths of different techniques (Bashar & Nayak, 2025). These models have proven particularly effective in detecting anomalies within complex, data-rich environments where traditional methods often fall short due to scalability or adaptability limitations.

## 2.4 AI in anomaly interpretation

Traditional anomaly detection methods often produce opaque outputs, making it difficult for users to interpret the results. Explainable Anomaly Detection (XAD**)** addresses the challenge of anomaly interpretation by providing clear, human-understandable explanations for why specific data points are flagged as anomalies (Z. Li et al., 2023a). XAD enhances user trust and supports more effective decision-making by revealing which features or patterns contributed to the anomaly and how these deviates from expected behaviour (Choi et al., 2022). This capability is especially critical in complex systems, where unexplained or poorly justified decisions can lead to operational errors or diminish stakeholder confidence.

XAD techniques can be categorized across several dimensions, each contributing to the interpretability of the detection process in distinct ways (figure 4). One key categorization is based on the placement of the explainability component within the detection pipeline. Pre-model techniques aim to enhance interpretability before the detection process, often through feature selection or data representation strategies. In-model techniques leverage inherently interpretable algorithms such as decision trees, rule-based systems, or probabilistic models that provide transparency during the detection phase (Y. Zhang, Li, et al., 2024). Post-model techniques, by contrast, interpret the output of complex or black-box models after anomalies have been identified, using approaches such as subspace analysis or surrogate modelling.

Another classification focuses on the scope of explanation. Global explanations aim to elucidate the overall decision-making logic of the anomaly detection model, helping stakeholders understand how it functions in general. Local explanations concentrate on specific flagged anomalies, offering insights into why particular data instances were considered anomaly. Additionally, XAD methods can be distinguished by their model dependency. Model-agnostic techniques are designed to work with any detection model, while model-specific techniques are tailored to particular algorithms and leverage their internal structures.

Explanations can also be differentiated by their methodological approach. *Feature-based* methods identify the contribution of individual input features to the anomaly score, while *sample-based* methods compare detected anomalies to reference or expected samples. Some advanced techniques integrate both approaches to provide more comprehensive and interpretable insights.

Several specific techniques have been developed to enable XAD. *Approximation-based* methods employ surrogate models to approximate and interpret the behaviour of detection algorithms. Perturbation-based techniques assess how changes in input data affect anomaly scores (Tang et al., 2024). Reconstruction error-based approaches, common in autoencoder architectures, highlight the input features most responsible for reconstruction failure, thereby indicating anomalies (Tsai & Jen, 2021). Gradient-based methods, often used in neural networks, analyse the influence of features via gradients. Causal inference-based methods go further by identifying causal relationships between variables and outcomes (M. Li et al., 2022). Finally, visualization-based methods support interpretability through interactive visual tools that help users intuitively explore and understand anomaly patterns (Talmoudi et al., 2021).

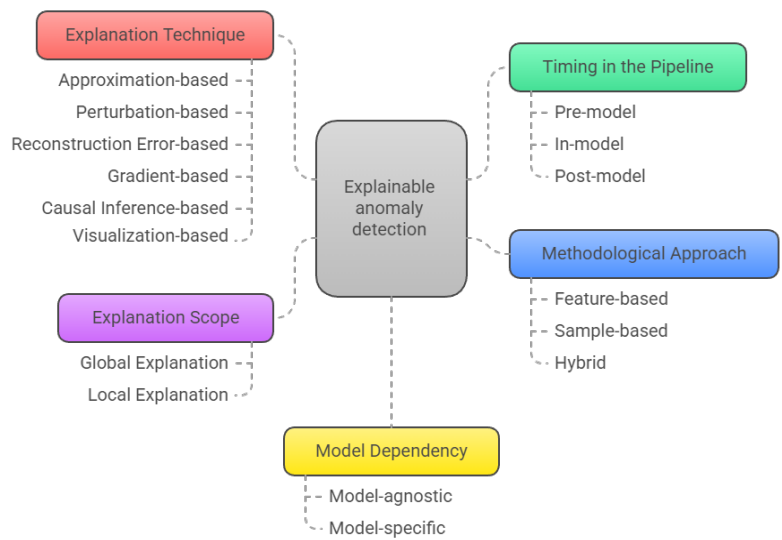

*Figure 4. XAD classification based on five criteria (Z. Li et al., 2023)*

More recently, Large Language Models (LLMs) have been investigated for their potential in anomaly explanation. One notable example is HuntGPT, which combines machine learning-based anomaly detection with explainable AI and LLMs to enhance cybersecurity operations(Ali & Kostakos, 2023). Its layered architecture—including an analytics engine, data storage, and user interface—is designed to improve trust, interpretability, and actionable response to detected threats. In other domains such as autonomous driving and robotic manipulation, LLMs have been employed for semantic anomaly detection, where visual inputs are translated into natural language descriptions. Structured prompts guide LLMs in identifying subtle or context-dependent anomalies that traditional methods often fail to detect (Y. Zhang, Cao, et al., 2024). These studies indicate that LLM-based explanations closely align with human reasoning, thereby improving the interpretability and accessibility of AI-driven systems.

**2.5 Limitations in Current Anomaly management approaches for complex systems**

Despite significant advancements in anomaly management for complex systems, several limitations persist that hinder broader applicability, robustness, and adaptability. These challenges are primarily rooted in the design philosophies and operational assumptions of current approaches, which often rely heavily on human decision-making. The dependence on human intervention limits the scalability and responsiveness of such systems, especially in complex and high-stakes environments where timely, autonomous actions are critical.

A major constraint in current anomaly management approaches is their reliance on rule-based modeling, which inherently lacks adaptability. Most existing methods for anomaly diagnosis and intervention are built upon predefined rules or heuristics derived from prior domain knowledge to define what constitutes "normal" system behavior (Nassif et al., 2021). Anomalies are then detected as deviations from these static definitions. While effective in narrowly scoped and controlled environments, this approach often fails in complex or open-ended contexts where system behaviors evolve over time (Shaukat et al., 2021). These systems are particularly ill-equipped to handle novel but valid patterns, often referred to as novelties, which may reflect legitimate adaptive behaviors rather than true anomalies. Moreover, sensor noise or hardware faults can further complicate detection by producing false positives. As a result, such rule-based approaches frequently require additional interpretive mechanisms to provide post-hoc explanations and support informed decision-making.

Another critical limitation lies in the inflexibility of current approaches for real-time decision-making and intervention. The exist, anomaly management approaches are frequently treated as discrete, sequential processes and human dependent: anomalies are first detected, explanations generated, and only then intervened. This introduces latency, making these approaches unsuitable for online or closed-loop complex systems, where timely response is essential. The delay between these steps can lead to missed opportunities or critical failures (Hao et al., 2023).

Moreover, current approaches generally lack the capacity to reconfigure their goals or dynamically adjust their operational boundaries. They are typically built for fixed objectives and static environments, rendering them ineffective in complex systems where tasks and priorities shift in response to internal feedback or external stimuli. Complex systems demand anomaly management systems that not only identify deviations but also adapt their decision-making frameworks as system goals evolve.

To overcome these challenges, future research in this context must focus on developing autonomous and adaptive anomaly management approaches capable of continuously learning from streaming data, updating internal representations based on environmental feedback, and accurately distinguishing between true anomalies, novelties, and noise. Crucially, these systems should also be capable of autonomously making decisions regarding appropriate interventions. Such approaches should be supported by flexible architectures that enable continuous adaptation, real-time goal reconfiguration, and seamless operation in complex environments. Additionally, these systems should retain the ability to incorporate expert guidance when necessary, allowing for human-in-the-loop interaction in cases that require domain-specific knowledge or ethical oversight.

## 3. In-Depth Exploration of Agentic AI

### 3.1 The Progression of AI

The first phase of AI is models that learn from data, primarily through supervised, unsupervised, and semi-supervised learning paradigms (Wu et al., 2025). In supervised learning, models are trained on labelled datasets to establish mappings between inputs and outputs using ground truth examples. This approach has been foundational in tasks such as image classification, speech recognition, and machine translation. In contrast, unsupervised learning operates without labelled data, aiming to uncover hidden patterns, clusters, or structures within the data using techniques such as k-means clustering, hierarchical clustering, and dimensionality reduction methods like Principal Component Analysis (PCA*)* and t-Distributed Stochastic Neighbour Embedding (t-SNE). Semi-supervised learning, positioned between these two extremes, leverages a small set of labelled data in combination with a larger pool of unlabelled data. It offers a practical solution in scenarios where manual annotation is costly or impractical (S. Li et al., 2024). Together, these traditional paradigms have laid the groundwork for modern AI research and remain integral to a wide range of applications.

AI agents mark the second major phase in the development of artificial intelligence. They first emerged in the 1990s as autonomous software programs designed to help users navigate the early, relatively simple internet. Today, an AI agent is regarded as an intelligent system or smart digital assistant (Boston Consulting Group, 2025). AI agents learn through interactions with an environment, receiving feedback in the form of rewards or penalties. This trial-and-error process enables AI agents to iteratively refine their strategies to maximize cumulative rewards (Kapoor et al., 2024). These agents are goal-driven (Boston Consulting Group, 2025) and possess the ability to retain knowledge across tasks, carry out complex functions, and independently determine when to access internal or external systems. This high level of autonomy enables them to operate efficiently, reduce human involvement, and streamline business processes (Boston Consulting Group, 2025). An example of this can be seen in a consumer goods company that used an AI agent to optimize its global marketing campaigns. Previously, the task required a team of six analysts each week; with the AI agent, one employee could complete it in under an hour. The agent automatically gathers marketing data, analyzes performance metrics, and generates optimization recommendations. After review and approval by a human operator, it implements the changes directly into media buying platforms (Boston Consulting Group, 2025).

AI agents rely on a set of integrated modules to function effectively. The Interface module manages communication with users, systems, and databases. The Memory module stores both short-term context and long-term knowledge from prior interactions. The Profile module defines the agent's goals, roles, and behavioral parameters. The Planning module uses AI models to create actionable strategies, while the Action module executes these plans through APIs. Together, these components enable AI agents to learn, adapt, and act intelligently within their environments (Boston Consulting Group, 2025).

The advent of deep learning marked a transformative leap in artificial intelligence, initiating the third stage of AI evolution. Deep neural networks—particularly convolutional neural networks (CNNs) and recurrent neural networks (RNNs)—have demonstrated the ability to learn hierarchical feature representations directly from raw data. By eliminating the reliance on manually engineered features, deep learning has driven major breakthroughs in fields such as computer vision, natural language processing, and speech recognition. Its success is largely attributed to its scalability with large datasets and the availability of high computational power.

One of the most significant advancements emerging from deep learning is the development of Large Language Models (LLMs) (Casella & Wang, 2025). These models are trained using self-supervised learning on massive text corpora and utilize transformer architectures to understand, generate, and manipulate natural language with impressive fluency and contextual depth. While fundamentally rooted in deep learning, LLMs are often fine-tuned using reinforcement learning from human feedback (RLHF), which helps align their outputs with human preferences and values. This hybrid training paradigm has greatly improved the performance, usability, and reliability of LLMs in real-world applications.

LLMs have demonstrated remarkable capabilities in zero-shot and few-shot learning across a wide range of natural language tasks. However, despite their success, they also present intrinsic limitations that scaling alone cannot resolve (Acharya et al., 2024). These challenges include the inability to access real-time information (A. Zhou et al., 2023), a tendency to hallucinate or generate inaccurate content (L. Wang et al., 2024), limited effectiveness in low-resource languages (Jadhav et al., 2024), weak performance in mathematical reasoning (Ahn et al., 2024), and a lack of temporal awareness (Chu et al., 2024).

The rise of large language models (LLMs) has renewed interest in AI agents and ushered in the fourth stage of AI, known as Agentic AI. This stage represents the integration of AI agents with the power of LLMs to interpret natural language instructions and autonomously determine when to activate tools such as web search, code execution, or data retrieval. This capability significantly enhances their operational effectiveness and unlocks new possibilities for dynamic, goal-oriented AI applications (Z. Shen, 2024). With the integration of memory systems, tool-use capabilities, and goal-directed prompting strategies, agentic AI systems are evolving into autonomous entities capable of interacting with other AI models, digital tools, and even real-world environments. Systems like GPT, BERT, PaLM, and LLaMA that can plan, reason, and act intelligently within complex, ever-changing contexts (Shavit et al., n.d.). Unlike traditional AI agents, which rely on passive input-output interactions, Agentic AI represents a shift toward proactive engagement with their surroundings (Rao Maka et al., 2021). This progression blurs the line between conventional predictive AI and autonomous, adaptive systems, positioning (Hughes et al., 2025).

## 3.2 Agent AI concept and comparison

The concept of "agenticness" describes the degree to which a system can operate autonomously and engage in sustained, context-sensitive behaviour (Shavit et al., n.d.). It involves the sophistication of goals a system can pursue, the complexity of the environments in which it operates, its adaptability to unfamiliar situations, and its capacity for independent execution (Liu & Yao, 2021). Agentic AI systems exemplify these traits, often operating with minimal supervision while coordinating complex actions across varied domains (Ge et al., 2023). They can shift objectives dynamically, adjust operational boundaries in real time, and maintain functional integrity even amid fluctuating environmental conditions. It moves beyond rule-based automation and static optimization. It can plan hierarchically, aligning broad strategic goals with low-level operational actions. They possess mechanisms for adaptive control, allowing them to respond effectively to new inputs, unanticipated challenges, or shifting task constraints (T. Wang et al., 2024). In many cases, these systems function not just as tools to support human work but as collaborative partners or fully independent agents capable of handling tasks in hazardous or inaccessible environments. Their ability to process natural language, interpret sensory inputs, and integrate multimodal information contributes to their effectiveness across diverse application areas—from cybersecurity and robotics to scientific research and industrial automation.

The heart of Agentic AI lies a powerful integration of deep learning particularly LLM with AI agents (Hu et al., 2023). While LLM endows these systems with rich representational capabilities and decision-making, AI agents provide a feedback-driven mechanism, allowing for control, adaptability, and the capacity to learn optimal behaviours through interaction with the environment (Bill & Eriksson, 2023). This synergy enables them to move beyond mere data processing and into the realm of autonomous action, where it can form internal models of the world, pursue long-term objectives, and revise strategies in real time. This is not simply an extension of earlier concepts such as AI agent but a reimagining of how they can be integrated to do planning, reasoning, show autonomous behaviour in complex, changing environments.

Traditional AI agents tend to operate within predefined, narrow scopes, often relying on static data classification or hardcoded responses. In contrast, Agentic AI is designed for adaptability, enabling AI agents to function across unstructured tasks, adjust to novel situations, and remain oriented toward high-level goals even when faced with uncertainty or conflicting objectives (Eigner & Händler, 2024). These systems are capable of continuous learning and improvement, guided by reinforcement mechanisms that help refine their decision-making over time or learn. The capacity to pursue (Xiang et al., 2024) complex objectives with minimal human intervention makes Agentic AI uniquely suited for real-world complex scenarios where adaptability, speed, and contextual awareness are essential. A comparison between AI agents and Agentic AI is provided in Table 1.

Table 1 compares the characteristics AI Agent with Agentic Ai.

| Characteristics | AI Agent | Agentic AI |
|---|---|---|
| Input Type | Single modality | Multi-modal |
| Interaction Mode | Passive Input-output | Proactive Interaction |
| Scope of Tasks | Narrow, Predefined | Broad, Unstructured |
| Adaptability | Low | High |
| Learning | Offline, Static | Continuous |
| Anatomy | Limited | High |
| Anomaly Explanation | Manual Interpretation | Context aware Interpretation |
| Anomaly Response | Manual Intervention | Autonomous Adaptive |
| Feedback Integration | None, Rule-based | Integrated Modules |
| System Architecture | Isolated, linear | Integrated and Recursive |
| Tool Use | No | Yes |
| Context Awareness | Minimal | High |

An illustrative example of Agentic AI in action is the Agent-S framework, which enables systems to interact with computer interfaces in ways that closely mimic human behaviour (Agashe et al., 2024). This framework incorporates hierarchical planning grounded in prior experience, updates its internal memory continuously, and uses a specialized agent-computer interface to achieve high-precision perception and control. By combining episodic memory, narrative context, and strategic reasoning, Agent-S demonstrates how Agentic AI can solve complex, interface-driven tasks in a closed-loop, autonomous fashion.

While agentic AI capabilities are transformative, they also introduce significant challenges. These systems may amplify the abilities of malicious actors, foster overreliance that weakens human oversight, or lead to delayed and diffuse consequences that are difficult to detect or mitigate. Interactions among multiple autonomous agents can also give rise to emergent behaviours that were neither predicted nor intended by their developers (Hexmoor et al., 2025). However, when properly coordinated, networks of specialized agents can collaborate efficiently distributing responsibilities and solving interdependent problems more effectively than any single system could on its own (G. Li et al., 2025). These systems capability to interface seamlessly with external tools—such as search engines, email systems, code interpreters, and databases—enhancing both their problem-solving capacity and operational scope.

## 4. AgenticAI in complex systems

Complex systems often involve a multitude of interacting entities, ranging from software agents and human operators to physical devices and sensors. Managing such complexity requires the seamless integration of diverse tools, machine learning models, and knowledge-based systems, each contributing specialized capabilities. These components must operate collaboratively, sharing information, coordinating tasks, and adapting to changes in real time. Moreover, in many systems, the goals of the system may shift in response to external conditions, stakeholder priorities, or emergent challenges. As a result, the system must be highly adaptive—not only in responding to environmental changes, but also in reconfiguring its strategies and objectives dynamically. Central to this adaptability is the crucial role of human operators, whose judgment and decision-making remain essential in guiding the system's overall direction and resolving complex or unforeseen issues. This layered and interconnected structure, adaptive coordination strategies, and scalable architectures that support continuous learning and autonomous reasoning across all participating elements.

AI agents and multi-agent architectures are widely used in managing complex systems due to their ability to learn and adapt through interaction with ever-changing environments. These AI agents serve as intelligent assistants that augment human decision-making, providing insights and recommendations that enhance the effectiveness and responsiveness of human operators. Unlike static decision-making models, AI agents continuously refine their behaviour based on feedback, enabling them to handle uncertainty, non-linearity, and evolving objectives. This makes them particularly well-suited for tasks such as autonomous control, resource allocation, and adaptive planning, where conditions and goals may shift unpredictably. By optimizing actions through trial-and-error, AI agents can develop policies that balance short-term performance with long-term adaptability, making them a powerful tool for maintaining stability and efficiency in dynamic, real-time systems.

Despite their adaptability, AI agents suffer from key limitations, primarily their reliance on trial-and-error learning. This approach often requires large amounts of data and time, making it inefficient and potentially risky in real-world or safety-critical applications. Similarly while effective in modelling decentralized and complex environments, face challenges in coordination, scalability, and stability. In multi-agent settings, emergent behaviours can be unpredictable, and agents may compete or interfere with one another, complicating convergence and control. These limitations highlight the need for more guided, data-efficient, and robust approaches in complex system management.

AI agents are generally not the best for anomaly management in complex systems due to their dependence on trial-and-error learning and long training cycles. Anomalies, by nature, are rare, unpredictable, and often high-stakes events that require immediate and accurate responses. Since AI agents learn optimal behaviour through repeated interactions and gradual reward optimization, they may fail to detect or respond appropriately to infrequent but critical anomalies. Additionally, exposing AI agents to anomalous scenarios during training is often impractical or unsafe, especially in domains like industrial automation, healthcare, or cybersecurity. Their lack of inherent interpretability and slow adaptation to unforeseen situations further limits their effectiveness in managing anomalies, where rapid, rule-based, or expert-informed interventions are often required.

Human oversight remains essential because these systems lack contextual understanding, intuition, and ethical judgment, which are often needed in high-stakes or ambiguous situations. Unlike AI agent , humans can quickly assess novel scenarios, incorporate real-time feedback, and make morally sound decisions without relying solely on historical data. Furthermore, AI agents may produce unpredictable or biased outcomes when faced with edge cases, requiring human intervention to ensure safety and accountability. While AI agent can assist in anomaly detection and interpretation, the final decision-making and intervention remained with human experts who can weigh risks, consider broader implications, and adapt to dynamic conditions beyond the system's training scope.

While LLMs demonstrate impressive linguistic capabilities, they remain fundamentally disconnected from real-world systems - lacking native integration with live data streams, sensor networks, domain-specific software, or external tools (Liang & Tong, 2025). This architectural isolation creates three critical limitations: (1) inability to perceive dynamic environmental states due to static training data, (2) absence of direct actuator control for physical/digital interventions, and (3) dependence on middleware for any time-sensitive operations. Current agentic AI research (Yuan et al., 2024) attempts to mitigate these through tool-augmented architectures where LLMs function as cognitive cores.

Within the literature, agentic AI models can be broadly categorized into two main groups based on their tool augmentation architecture: single-tool and multi-tool systems. Single-tool approaches involve the integration of one external API or utility into the LLM's reasoning framework to enhance performance on defined tasks. Examples include LLaMA (PAL) and MathPrompt, which employ deterministic code execution for mathematical problem-solving, while Program-of-Thought adapts similar methodologies for table-based question answering with guaranteed syntactic validity (Bi et al., 2024). Code4Struct extends this paradigm to structured prediction tasks like event extraction through constrained decoding (X. Wang et al., 2023), demonstrating how domain-specific tool specialization can outperform general-purpose LLMs by 12-18% on benchmark evaluations.

The transition to multi-tool systems introduces architectural innovations in tool orchestration. ToolFormer (Schick et al., 2023) pioneers learned tool invocation through API-aware fine-tuning, achieving 83% precision in autonomous tool selection across 7 utilities. GraphToolFormer (J. Zhang, 2023) advances this by combining graph reasoning modules with human-curated toolchains, improving relational reasoning F1 scores by 29% on web questions. TALM (Tool-Augmented Language Models) demonstrates how iterative tool refinement can reduce hallucination rates by 40% in knowledge-intensive tasks (Parisi et al., 2022). These systems reveal an emerging design trade off: while single-tool integrations offer deterministic correctness (e.g., 100% accurate calculator outputs), multi-tool ecosystems enable compositional reasoning at the cost of increased latency (~300ms/tool call) and coordination complexity (Zhou et al., 2023).

Multi-tool agentic AI systems are designed to handle complex, often multi-modal workflows requiring dynamic orchestration of diverse specialized tools. These frameworks excel in domains demanding web interaction (WebGPT; Nakano et al., 2021), visual analysis (MM-ReAct), code generation (HuggingGPT;(Y. Shen et al., 2023)), and real-time data retrieval (GeneGPT; (Jin et al., 2024)), where they demonstrate three key architectural advantages: (1) cross-domain tool composition (e.g., Chameleon's 12-tool integration for multimodal tasks), (2) hierarchical task decomposition (as in HuggingGPT's LLM-mediated subtask routing), and (3) output verification (exemplified by WebGPT's reference grounding).

The paradigm positions LLMs as meta-reasoners that parse objectives (ReAct's 'Thought-Action-Observation' loops), delegate to domain tools (ART's 8-tool ensemble for BigBench; Suzgun et al., 2022), and synthesize outputs (Visual ChatGPT's vision-language fusion). GeneGPT's integration with 38 NCBI APIs shows how tool multiplicity enables biomedical precision (87.4% accuracy on genomics QA), while MM-ReAct proves multimodal systems can achieve 23% higher success rates in vision-language tasks versus unimodal baselines. However, these systems incur nontrivial latency costs (~2-5× slower than single-tool approaches) and require sophisticated tool-embedding techniques to maintain context across heterogeneous interfaces (Shaukat et al., 2021).

Agentic AI systems are increasingly capable of assuming human decision-making responsibilities in complex systems. These systems may reduce cognitive load, improve response times, and handle multivariate optimization beyond human capability. However, fundamental challenges persist in temporal synchronization (latency between perception and action), verifiable reliability (especially in safety-critical domains), and maintaining consistency between the model's parametric knowledge and external system states. Such limitations—rooted in the brittleness of tool-augmented LLMs when faced with novelty, the opacity of their failure modes, and their inability to contextualize decisions ethically— necessitate human oversight. Thus, while agentic AI can augment human judgment in structured tasks (e.g., triaging routine cases), it cannot yet internalize the holistic reasoning, moral agency, or adaptive intuition required for fully autonomous operation in open-world systems.

## 4.1 Maritime shipping anomaly management

To date, only a limited number of use cases have explored the deployment of Agentic AI for complex system's anomaly management. One notable case involves the application of agentic AI system for anomaly diagnosis, decision-making, and planning in maritime shipping asset management and maintenance (Zhuang, Yuchen et al., 2023). Anomaly management in this domain is particularly challenging due to complex operating conditions, fluctuating loads, complex interdependencies among subsystems, and limited availability of labelled training data. High false positive rates can lead to increased maintenance costs and diminish confidence in condition-based monitoring. Moreover, interpreting anomaly detection outputs often requires human judgment, as anomalies can arise simultaneously in multiple data streams and may originate from environmental factors or distant subsystems before propagating downstream.

In response to these challenges, (Zhuang, Yuchen et al., 2023) developed an agentic AI system that integrates reasoning and planning. This system decomposes tasks into subtasks and coordinates the execution of these subtasks using a suite of specialized tools. A central component of the system is a domain-specific knowledge graph (KG) that represents relationships between ship components, operational parameters, and environmental conditions. This KG serves as a semantic layer that contextualizes real-time sensor data, technical specifications, and performance benchmarks. By bridging raw operational data with high-level decision-making processes, the KG helps the system identify the relevant components and data sources needed to monitor ship performance and detect anomalies.

The agentic AI system utilizes the reasoning capabilities of an LLM not only to interpret user queries but also to infer potential sources of anomalies based on available data. Given that the causal pathways for anomaly propagation are not always explicitly encoded in the knowledge graph, the LLM plays a critical role in generating hypotheses about the origin and spread of faults (R. Wang et al., 2023). Additionally, an "LLM-as-a-judge" module was implemented to evaluate the appropriateness of tool use. This module reviews the sequence of tool invocations and determines whether the agent's actions align with the user's objectives or if further tool interactions are required.

The system also allows for the exploration of anomalies from various user perspectives or personas, enabling role-specific analysis and decision-making. By integrating real-time data, contextual information, and reasoning capabilities, the system provides a comprehensive view of ship operations and facilitates informed decision-making (Timms & Langbridge, n.d.-a) . The results of the case study suggest that such an agentic AI system can effectively resolve user queries by identifying relevant subsystems, evaluating sensor data, and reasoning through potential operational impacts. Particular attention was given to the system's capacity for context resolution and intelligent tool selection, which proved essential in accurately diagnosing anomalies and supporting proactive maintenance decisions.

This case study underscores the broader potential of Agentic AI to serve as a unifying paradigm for reasoning, perception, and action in complex systems. By combining LLM-driven reasoning with structured domain knowledge and interactive tool use, these systems exemplify a new generation of autonomous systems capable of managing anomaly or the complex systems and executing high-stakes tasks with minimal human oversight.

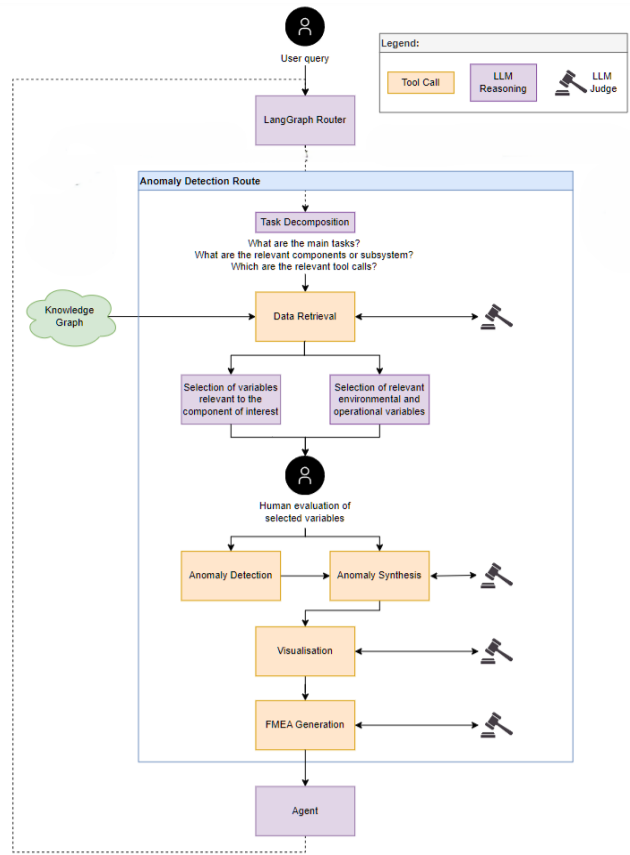

Figure 5 Agentic AI system for Maritime shipping anomaly management (Timms & Langbridge, n.d)

**4.2 Darktrace's Immune System as an Anomaly in Intrusion Detection and Prevention**

Traditional Intrusion Detection and Prevention Systems (IDPS) rely on predefined rules or signatures to monitor networks for known malicious activity, log incidents, and respond through automated blocking or human intervention. In contrast, Darktrace's Enterprise Immune System represents a significant departure from this conventional model by employing agentic AI—an autonomous, self-learning framework that approaches cybersecurity as a dynamic, adaptive challenge rather than a static, rule-based process. Whereas traditional IDPS struggle to detect novel attack vectors (e.g., zero-day exploits, polymorphic malware), Darktrace's system leverages continuous behavioural learning to establish dynamic baselines for all network entities, including devices, users, and applications. By analysing real-time deviations—such as anomalous data transfers, lateral movement, or privilege escalation—the system autonomously enforces countermeasures (e.g., quarantining compromised devices, throttling suspicious connections) without requiring human intervention. This agentic AI-driven response addresses a fundamental limitation of conventional IDPS: their dependence on historical threat data, which inherently renders them reactive rather than proactive (Castellanos, 2021).

However, the shift toward fully autonomous threat mitigation introduces several challenges. While Darktrace's adaptive learning capabilities reduce false positives over time (Weigand, 2025), its agentic nature raises critical concerns regarding accountability—such as erroneous actions that may disrupt legitimate operations—and the risks of overreliance on AI in high-stakes scenarios (e.g., sophisticated nation-state attacks where human judgment remains essential). Additionally, the system's opaque decision-making processes complicate regulatory compliance and auditability, presenting hurdles for organizations operating under strict governance frameworks.

Despite these challenges, the benefits of agentic AI in cybersecurity are substantial. By automating threat detection and response, Darktrace significantly reduces incident response times and operational costs while enhancing overall system resilience (Castellanos, 2021; Weigand, 2025). This automation streamlines security workflows, minimizes manual intervention (Bokkena, 2024), and improves Security Operations Centre (SOC) efficiency by enabling faster, data-driven decision-making (Columbus, 2025). Industry research supports this trend: Gartner predicts that AI will enhance SOC efficiency by 40% by 2026, fundamentally transforming cybersecurity roles toward AI oversight, maintenance, and strategic security management (Mikhailov, 2023). Thus, while Darktrace's Enterprise Immune System exemplifies the potential of agentic AI to revolutionize cybersecurity, its broader adoption necessitates careful consideration of ethical, operational, and regulatory implications to ensure responsible and effective deployment.

## 5. Dissuasion

Let us start this section with the research questions that we wanted to address. Regarding RQ1, AI based anomaly diagnose approaches leverage unsupervised learning and deep learning techniques to achieve remarkable accuracy in identifying anomaly from normal patterns. Unsupervised methods such as autoencoders, isolation forests, and clustering algorithms excel at detecting anomalies without labelled training data by modelling complex data distributions. Meanwhile, deep learning approaches—including recurrent neural networks (RNNs) for temporal data and graph neural networks (GNNs) for relational anomalies—can uncover subtle, nonlinear patterns indicative of sophisticated threats. These systems not only flag anomalies but increasingly provide interpretable explanations (e.g., via SHAP values or attention mechanisms), aiding human analysts in diagnosing root causes. Despite these advances in detection and interpretation, autonomous intervention remains a persistent challenge. Most anomaly management systems still depend on human analysts to evaluate and act upon alerts, creating a critical disconnect between diageneses and resolution. As to RQ2, current anomaly diagnosis in complex systems faces significant limitations due to the interconnected yet heterogeneous nature of subsystems, each operating with distinct goals while contributing to an overarching system objective. This heterogeneity creates challenges in anomaly attribution, as a detected anomaly in one component may propagate cascading effects across other entities, complicating root-cause analysis. Additionally, inconsistencies in data formats, conflicting operational priorities among subsystems, and lack of standardized cross-system telemetry further obscure anomaly interpretation. These technical disparities are often compounded by organizational silos, where disparate teams responsible for different subsystems may disagree on anomaly severity or mitigation strategies. Consequently, anomaly management in these systems remains heavily reliant on human expertise, requiring coordinated intervention from domain specialists with divergent perspectives—a process that is slow, prone to misalignment, and ill-suited for real-time response in complex environments. Regarding RQ3, Agentic AI fundamentally differs from AI agent through its advanced autonomy, dynamic adaptability, and strategic long-term goal management. Unlike traditional AI agent systems that operate within predefined rules and require explicit human programming for each scenario, Agentic AI exhibits goal-directed agency, enabling it to independently interpret objectives, dynamically adjust strategies in real-time, and make context-aware decisions without constant human oversight. This is achieved through three core capabilities: (1) Autonomous decision-making, where systems like Darktrace's Antigena can quarantine devices or block network traffic without human approval by continuously evaluating risk thresholds; (2) Meta-adaptability, allowing agents to not just learn from data but also evolve their own learning protocols—for instance, OpenAI's GPT-4 can now self-refine its reasoning paths through recursive self-improvement techniques; and (3) Multi-horizon goal management, demonstrated by systems like DeepMind's AlphaFold which balances immediate protein structure predictions with long-term biomedical research impacts. To RQ4, the integration of Agentic AI with LLMs and AI agents can significantly enhance the capabilities of real-time anomaly detection, interpretation, and intervention in complex systems. Agentic AI enable autonomous agents to operate with goal-directed behaviour, planning, and adaptive decision-making, while LLMs contribute advanced semantic understanding, contextual reasoning, and natural language interaction. In this integrated agent, LLMs can be employed to interpret sensor data streams, logs, and system messages in real time, identifying subtle patterns or deviations that may indicate anomalies. These interpretations are then used by AI agents to assess the severity and potential impact of the anomaly, leveraging domain-specific knowledge, real time and historical data and using different specific tools. The agents can autonomously determine and execute appropriate interventions—such as reconfiguring system parameters, alerting human operators with natural language explanations, or initiating fail-safe protocols—based on predefined objectives and real-time constraints. This synergy not only improves the accuracy and timeliness of anomaly detection but also enhances system resilience by enabling proactive and context-aware responses in dynamic and uncertain environments.

## 5.1 Safety, Complexity, and Contextual Awareness

In safety-critical domains, the stakes of anomaly detection are extraordinarily high. False negatives may result in catastrophic system failure, while false positives lead to unnecessary interventions, lost time, and eroded trust in automation. Agentic AI can improve safety not only through higher diagnostic accuracy, but by fostering system-level understanding. These systems interpret anomalies not just as data points, but as contextual events embedded within operational narratives. They consider how, where, and why a deviation occurred, and what downstream effects it may trigger.

For instance, in the maritime anomaly detection system can leverages real-time sensor streams, system performance metrics, environmental conditions, and a knowledge of entities and stockholders and interactions. This allows the system to distinguish between benign deviations and critical anomalies (e.g., failures in the propulsion or navigation subsystems). More importantly, it enables explanatory and anticipatory modelling, allowing operators to trace potential root causes, forecast the evolution of faults, and execute preventive maintenance protocols.

Such contextual awareness is critical for responding to ad hoc events, including novel faults, rare edge cases, or multi-causal disruptions. In conventional systems, these events often go undetected due to their deviation from known data patterns. However, an Agentic AI system, equipped with reasoning and analogical inference, can generalize from limited evidence, hypothesize potential causes, and query relevant tools—such as simulation environments or diagnostic submodules—on the fly. This ensures robustness under uncertainty, a critical requirement for complex system management.

## 5.2. Agentic AI as Collaborative Infrastructure

Another potential contribution of Agentic AI can be its role in human-AI collaboration in anomaly management. Far from replacing human oversight, these systems can act as intelligent partners, enabling multi-persona interactions tailored to the specific roles and responsibilities of engineers, analysts, technicians, or managers. By translating low-level telemetry into high-level summaries, and providing rationales for actions and alerts, Agentic AI can support faster decision cycles and more informed interventions.

The LLM-as-a-judge paradigm where the agent evaluates whether selected tools and actions align with the user's query and system goals introduces a new form of intent-aligned intelligence. It minimizes errors caused by tool misuse or misinterpretation, while continuously refining strategies based on task outcomes. This approach is particularly impactful in complex systems, where actions taken in one subsystem (e.g., fuel efficiency optimization) may influence others (e.g., cooling, power distribution).

Moreover, knowledge integration functions as a semantic foundation for cross-tool interoperability. It bridges complex system operations with digital diagnostics by formalizing relationships between subsystems, setpoints, thresholds, and failure types. It also captures temporal and causal structures, enabling the agent to infer how anomalies propagate and where interventions would be most effective. This type of systemic reasoning is essential for managing multi-modal complexity, especially in infrastructure that spans physical, digital, and cyber-physical layers.

## 5.3. Computational and Ethical Considerations

Despite these advancements, deploying Agentic AI in real-world settings is not without challenges. The integration of planning, deep learning, and tool execution significantly increases computational overhead. Fine-tuning LLMs for domain specificity, aligning outputs with human expectations, and managing feedback loops between tools and models require extensive engineering. More critically, transparency, interpretability, and accountability remain open research areas. In safety-critical domains, it is essential that system outputs are traceable, auditable, and fail-safe particularly when used for autonomous decision-making without human intervention.

There are also ethical concerns about over-reliance, displacement of human operators, and the risk of correlated failure modes, where multiple agentic systems, trained under similar biases or conditions, fail in concert. This raises the need for robust simulation environments, interdisciplinary and regulatory frameworks to ensure responsible deployment.

## 5.4 Toward the Future of Autonomous Intelligence in complex Systems

Agentic AI is more than an extension of AI agent it is a holistic redesign of how intelligence is embedded into complex systems. Its strengths lie in its generative flexibility, tool-use adaptability, and contextual sensitivity. These capabilities are not confined to anomaly detection alone but extend to planning, control, coordination, and even collaborative autonomy among agents. The ability to break down goals, choose appropriate tools, and generate explainable plans positions Agentic AI as the nervous system of future intelligent in complex systems.

# 6. Conclusion

This paper has examined the transformative potential of Agentic AI in the autonomous management of anomalies within complex systems. As these systems grow increasingly dynamic, distributed, and interdependent, traditional human-centred approaches to anomaly detection and response have become inadequate in addressing the speed, scale, and complexity of complex systems. By integrating the autonomous reasoning and goal-directed behaviour of Agentic AI with the semantic depth and contextual comprehension of LLMs a new paradigm emerges—one capable of continuously interpreting data, adapting to evolving conditions, and executing precise interventions with minimal human oversight. Agentic AI distinguishes itself from AI agent through its ability to learn from heterogeneous data sources, synthesize cross-disciplinary insights, and respond to anomalies proactively and autonomously. This capability not only enhances detection accuracy and response speed but also increases the resilience and adaptability of complex systems operating under uncertainty. The findings of this study highlight the significant limitations of conventional anomaly management approaches and make a compelling case for the transition toward intelligent, self-managing systems powered by Agentic AI. Future research should further explore the integration of Agentic AI with digital twin technology, real-time simulation environments, and human-in-the-loop architectures to ensure transparency, trust, and safety in high-stakes domains. As Agentic AI continues to evolve, it holds the promise of redefining how complex systems are monitored, understood, and controlled—shifting the role of human operators from reactive problem-solvers to strategic supervisors in an ecosystem of intelligent, autonomous agents.

Author contributions: Reza Vatankhah Barenji: writing—review & editing, supervision, project administration, conceptualization, validation, methodology, investigation, formal analysis, data curation. Sina Khoshgoftar: writing—review & editing, writing—original draft, validation, methodology, investigation, formal analysis, data curation.

Funding:   No funding supported for this work

Data availability:  No datasets were generated or analysed during the current study

313131